\documentclass[11pt]{article}

\usepackage[preprint]{acl}
\usepackage{times}
\usepackage{latexsym}
\usepackage{stmaryrd}
\usepackage{bbm}
\usepackage{amsfonts}
\usepackage{multirow}
\usepackage{amsmath} 
\usepackage{arydshln}
\usepackage{tabularx,booktabs,makecell,array}
\usepackage{booktabs}
\usepackage[most]{tcolorbox}
\usepackage{enumitem}
\usepackage{cleveref}
\usepackage[table]{xcolor}

\definecolor{modelpink}{HTML}{F7EDEA}   
\definecolor{modellav}{HTML}{F3EFE4}    
\definecolor{modelmint}{HTML}{EDF5EC}   
\definecolor{modelcyan}{HTML}{EAF3F7}   

\definecolor{blockgray}{gray}{0.7}
\usepackage[T1]{fontenc}

\usepackage[utf8]{inputenc}

\usepackage{microtype}

\usepackage{inconsolata}

\usepackage{graphicx}

\title{Beyond Unimodal Shortcuts: MLLMs as Cross-Modal Reasoners for Grounded Named Entity Recognition}

\author{
 \textbf{Jinlong Ma\textsuperscript{1}},
 \textbf{Yu Zhang\textsuperscript{1}},
 \textbf{Xuefeng Bai\textsuperscript{1}},
 \textbf{Kehai Chen\textsuperscript{1}},
 \textbf{Yuwei Wang\textsuperscript{2}},\\
 \textbf{Zeming Liu\textsuperscript{3}},
 \textbf{Jun Yu\textsuperscript{1}},
 \textbf{Min Zhang \textsuperscript{1}}
\\
\\
\\
 \textsuperscript{1}Harbin Institute of Technology, Shenzhen, China,\\
 \textsuperscript{2}Institute of Computing Technology Chinese Academy of Sciences,\\
 \textsuperscript{3}Beijing University of Aeronautics and Astronautics
\\
 \small{
   \textbf{Correspondence:} \href{chenkehai@hit.edu.cn}{chenkehai@hit.edu.cn}
 }
}

\begin{document}
\maketitle
\begin{abstract}
Grounded Multimodal Named Entity Recognition (GMNER) aims to extract text-based entities, assign them semantic categories, and ground them to corresponding visual regions. 
In this work, we explore the potential of Multimodal Large Language Models (MLLMs) to perform GMNER in an end-to-end manner, moving beyond their typical role as auxiliary tools within cascaded pipelines.
Crucially, our investigation reveals a fundamental challenge: MLLMs exhibit \textit{modality bias}, including visual bias and textual bias, which stems from their tendency to take unimodal shortcuts rather than rigorous cross-modal verification.
To address this, we propose Modality-aware Consistency Reasoning (\textbf{MCR}), which enforces structured cross-modal reasoning through Multi-style Reasoning Schema Injection (MRSI) and Constraint-guided Verifiable Optimization (CVO). 
MRSI transforms abstract constraints into executable reasoning chains, while CVO empowers the model to dynamically align its reasoning trajectories with Group Relative Policy Optimization (GRPO).
Experiments on GMNER and visual grounding tasks demonstrate that MCR effectively mitigates modality bias and achieves superior performance compared to existing baselines.
\begingroup
\renewcommand{\thefootnote}{}
\footnotetext{The code and data are released at \url{https://github.com/aaaalonga/MCR}.}
\addtocounter{footnote}{-1}
\endgroup
\end{abstract}
 \section{Introduction}
\label{sec:intro}
Grounded Multimodal Named Entity Recognition (GMNER, \citealp{gmner}) aims to organize key multimodal information into structured representations, which simultaneously identifies named entities in the text and grounds them to their corresponding visual bounding boxes. 
As a foundational task, GMNER facilitates various downstream applications, such as recommendation systems~\cite{acharya2023llm} and knowledge-based question answering~\cite{zhang-etal-2024-question,sun2023think}.

Recently, Multimodal Large Language Models (MLLMs, ~\citealp{qwen2_5vl, mimo}) have achieved remarkable performance on various vision-language tasks.
This progress has motivated researchers to explore their use for GMNER~\cite{tang2025unco,tang2025refineg}. 
However, these approaches typically employ MLLMs as auxiliary tools like image descriptors within cascaded pipelines, which inevitably introduce cumulative error propagation~\cite{multigrained} and incur additional computational costs~\cite{scanner}.
\begin{figure}[t]
  \centering
   \includegraphics[width=1.0\linewidth]{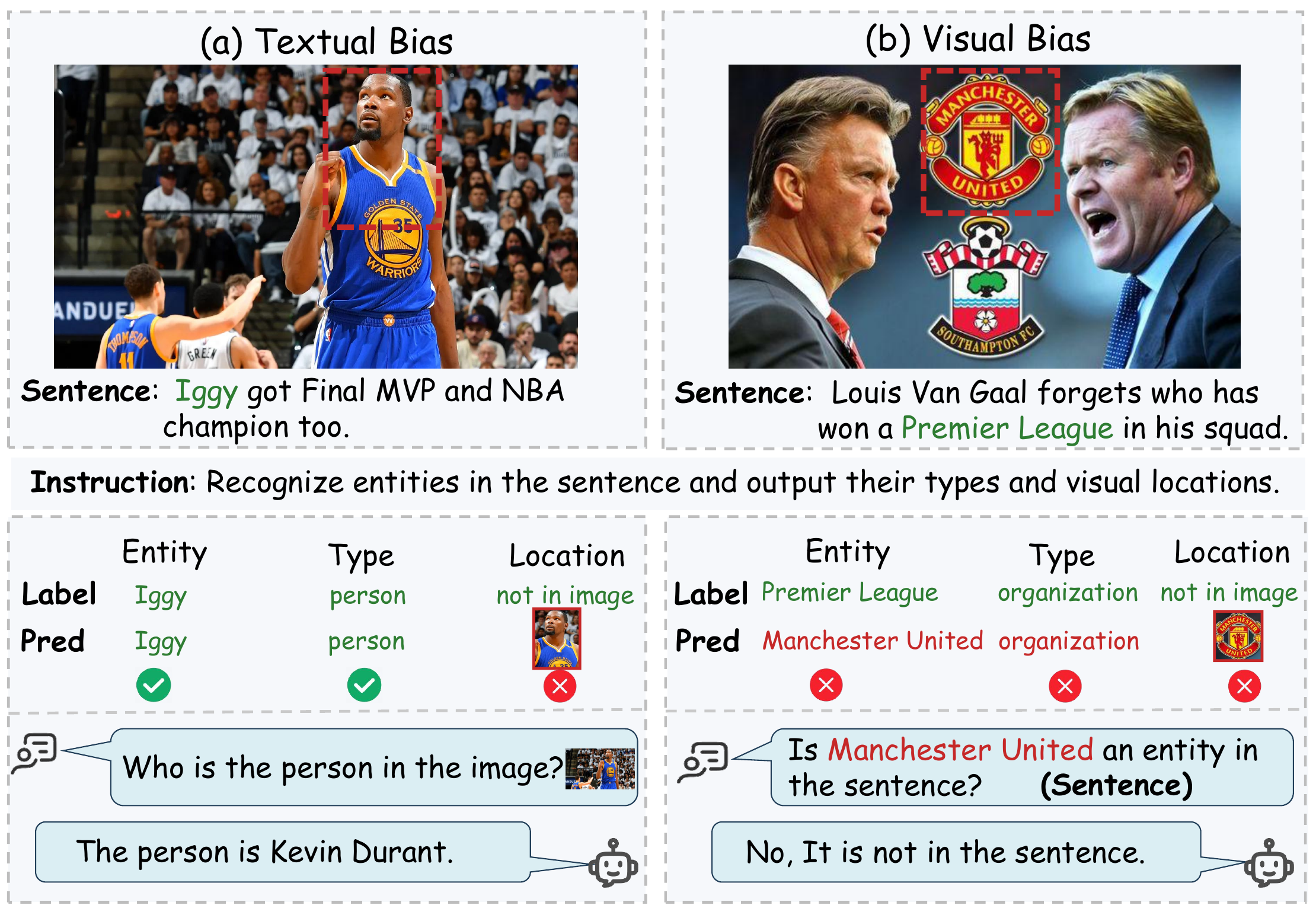}
   \caption{Error patterns caused by modality bias in GMNER due to the model's tendency to hallucinate correlations based on unimodal heuristics rather than rigorous cross-modal verification. 
   }
   \label{fig:modality_preference}
\end{figure}

In this work, we take the first step toward exploring the potential of MLLMs for end-to-end GMNER by reformulating it as a generative reasoning task.
Our investigation reveals that direct application of MLLMs to GMNER faces a critical pathology: \textit{modality bias}, characterized by the model's tendency to hallucinate correlations based on unimodal heuristics rather than rigorous cross-modal verification.
As shown in Figure~\ref{fig:modality_preference} (a), textual bias causes the model to disregard visual evidence: despite correctly recognizing ``Kevin Durant'' with image-only input, it incorrectly grounds the text-only entity ``Iggy'' to the bounding box of ``Kevin Durant''. 
Symmetrically, visual bias leads to the neglect of textual semantics.
In Figure~\ref{fig:modality_preference} (b), the model overrides textual context, erroneously recalling ``Manchester United'' as a named entity driven by visual cues, ignoring its absence in the textual context. 
We further conduct quantitative analyses that empirically confirm the severity and prevalence of modality bias for different MLLMs in Table~\ref{tab:n_rate}.
These reveal that MLLMs are prone to taking cognitive shortcuts rather than engaging in rigorous deduction, required for strict cross-modal grounding.

To address this, we propose \textbf{M}odality-aware \textbf{C}onsistency \textbf{R}easoning (\textbf{MCR}), which enforces structured cross-modal reasoning to mitigate modality bias through Multi-style Reasoning Schema Injection (MRSI) and Constraint-guided Verifiable Optimization (CVO). 
Specifically, MRSI transforms abstract constraints into executable reasoning chains by synthesizing and injecting diverse reasoning templates to explicitly model the structural dependencies.
Furthermore, to empower the model to autonomously explore reasoning trajectories within these structural bounds, CVO is proposed to dynamically align intermediate reasoning process with Group Relative Policy Optimization (GRPO)~\cite{deepseek}.
This optimization mechanism punishes unimodal shortcuts and encourages the model to generate constraint-faithful rationales, effectively rectifying the intrinsic modality bias.
Extensive experiments on Multimodal Named Entity Recognition~\cite{mner_mi,gmner} and Visual Grounding~\cite{grec} benchmarks verify that our method, applied to Qwen2.5-VL and Mimo-VL, achieves superior performance compared to existing baselines.
In-depth analyses confirm that our design explicitly facilitates cross-modal reasoning, effectively mitigating modality bias.

In summary, our contributions are as follows:
\begin{itemize}[itemsep=2pt,topsep=0pt,parsep=0pt,leftmargin=11pt]
    \item We identify \textit{modality bias} in MLLM-based end-to-end GMNER, revealing that models are prone to unimodal cognitive shortcuts.
    \item We propose a MCR framework, which enforces explicit, constraint-faithful reasoning through schema injection and verifiable optimization against modality bias.
    \item We achieve superior performance on multiple benchmarks, demonstrating that structured reasoning is essential for precise cross-modal grounding.
\end{itemize}

\section{Related Work}
\label{sec:related}

\subsection{Multimodal Named Entity Recognition (MNER)}
MNER~\cite{mner} extracts and classifies named entities from image-text pairs. As a fine-grained extension, Grounded MNER (GMNER)~\cite{gmner} requires the model to simultaneously recognize the named entities and localize visually present entities via bounding boxes.
Existing studies primarily focus on refining cross-modal alignment to suppress visual noise~\cite{liu2024hierarchical,bao2024contrastive} and enhancing generalization for unseen entities~\cite{granular_mapper}. 
With the emergence of Multimodal Large Language Models (MLLMs)~\cite{qwen2_5vl,mimo}, recent studies~\cite{tang2025refineg,tang2025unco,scanner} have started to integrate them into GMNER, a pivotal prerequisite for constructing knowledge graphs~\cite{zhong2023comprehensive,li2020real,li2024llm} and facilitating knowledge graph question answering~\cite{xu2025memory,sun2023think}.
These approaches primarily exploit the vast semantic priors inherent in MLLMs to refine and align multimodal feature representations, thereby facilitating more accurate entity-image association.

In this work, we move beyond feature alignment to fully exploit the cross-modal reasoning potential of MLLMs, enabling a holistic multimodal interplay for rigorous consistency verification.

\subsection{Reasoning in MLLMs}
MLLMs have achieved remarkable success across a wide range of domains~\cite{zhu-etal-2025-benchmarking,qwen2_5vl,Zhang_2025_ICCV,zheng2025locot2v,zuo-etal-2025-inimagetrans}, demonstrating exceptional capabilities in integrating and reasoning over heterogeneous data.
Recent advancements in MLLMs have catalyzed a paradigm shift in complex reasoning tasks~\cite{li2025system,chen2025towards,kumar2025llm，weifirst}. 
By introducing explicit reasoning processes into language-level Chains of Thought (CoT), these models decompose intricate problems into granular, sequential sub-steps~\cite{wei2022chain,wang2022self,gao2023pal}. 
This reasoning-centric paradigm has proven instrumental in mitigating hallucinations arising from modality misalignment~\cite{multimodal_cot,li2025imagine,zhang2025reasongen}, significantly enhancing both the accuracy and stability of multi-step inference.

\subsection{Modality bias in MLLMs}
Recent works identify \textit{modality bias}~\cite{debiasing,zhang2026instructionanchorsdissectingcausal,leng2024curse,zhang2025evaluating} in MLLMs, observing that models often exhibit intrinsic inclinations toward specific modalities.
To address this, prevailing strategies typically employ Reinforcement Learning from Human Feedback (RLHF) by curating extensive preference datasets~\cite{ouyang2022training,wang2024mdpo} to enable MLLMs to distinguish between hallucinated and grounded content, effectively mitigating bias and hallucinations.

In this work, we attribute modality bias in GMNER to cognitive shortcuts, where models bypass rigorous verification in favor of unimodal heuristics. 
To rectify this, we propose Modality-aware Consistency Reasoning (MCR) to explicitly model the interplay between modalities to verify entity existence and spatial alignment, thereby enforcing rigorous cross-modal consistency.
\begin{figure*}[ht]
  \centering
  \hfill
  \includegraphics[width=1.0\linewidth]{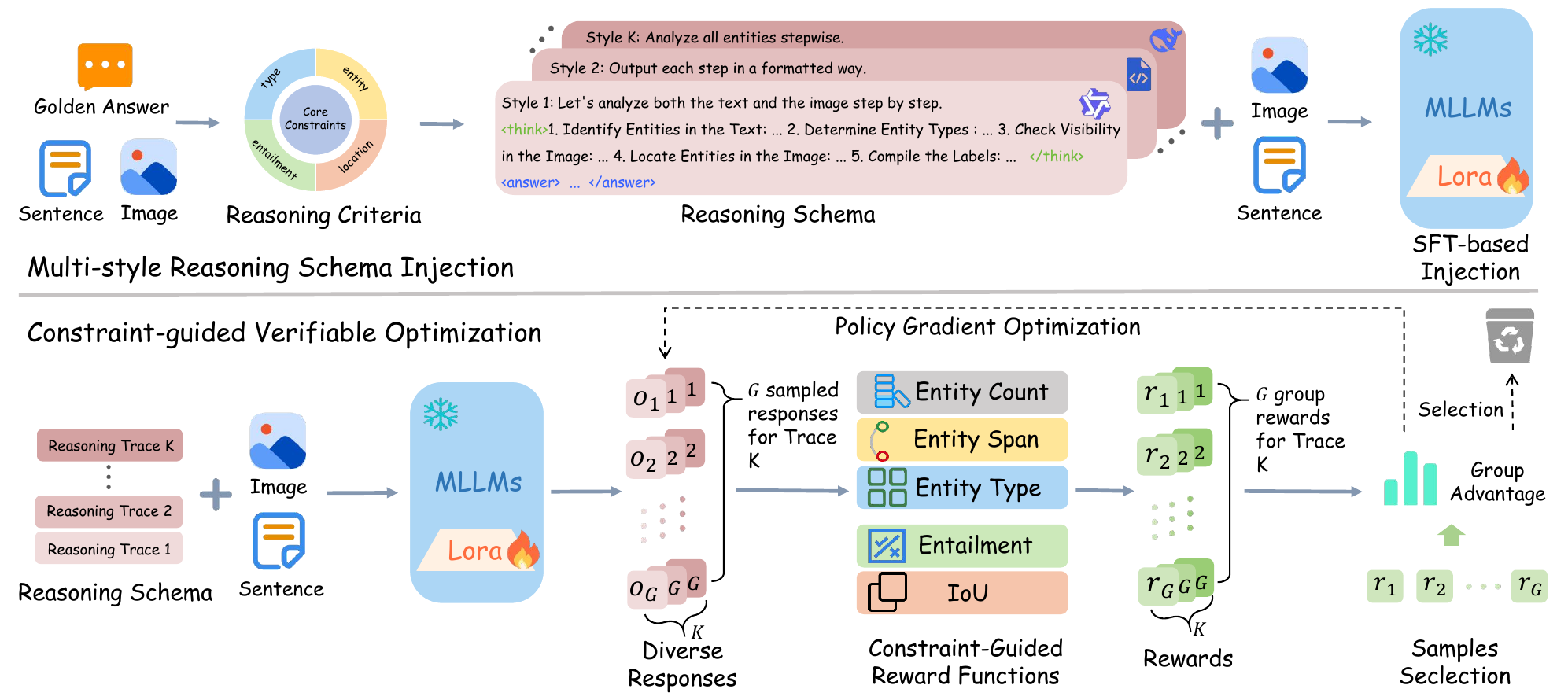}
  
  \caption{\textbf{The Framework of MCR.} The framework consists of two stages: (1) Multi-style Reasoning Schema Injection constructs diverse reasoning schema $\mathcal{D}_{\mathcal{R}}$ by treating the core constraints as reasoning criteria and generating multiple reasoning styles from templates, LLMs, and MLLMs based on the image–text inputs and labels. A subset of $\mathcal{D}_{\mathcal{R}}$ is injected into MLLMs through supervised fine-tuning. (2) Constraint-guided Verifiable Optimization uses the remaining of $\mathcal{D}_{\mathcal{R}}$ and optimizes the model with verifiable reward functions derived from the core constraints, together with the GRPO algorithm, to enhance cross-modal consistency reasoning.}
  \label{fig:method}
\end{figure*}

\section{Task Formulation}
\label{sec:task}
Given a sentence $s$ and its associated image $v$, Grounded Multimodal Named Entity Recognition (GMNER) can be decomposed into two subtasks:
\paragraph{Multimodal Named Entity Recognition (MNER).} MNER recognizes entities in $s$ and assigns each entity a predefined type. And it produces pairs $(e_i, t_i)$, where $e_i$ is an entity span in $s$ and $t_i$ denotes its corresponding type. 
\paragraph{Entity Extraction \& Grounding (EEG).} EEG is parallels generalized Visual Grounding (VG). For each textual entity $e_i$, decide whether it is visually present in $v$. If present, output its bounding box $b_i$; otherwise, output $\text{None}$.
Accordingly, the GMNER output can be formulated as:
\begin{equation}
\mathcal{Y} =
\{(e_i,\ t_i,\ l_i) \}_{i=1}^{k_1},
\label{eq:output_format}
\end{equation}
where ${k_1}$ indicate the numbers of output triples in a sample, and $l_i$ is formed as:
\begin{equation}
l_i =
\begin{cases}
b_i=(x_1,y_1,x_2,y_2),& e_i \ \text{is grounded},\\[2pt]
\text{None},& e_i \ \text{is ungrounded},
\end{cases}
\label{eq:location_format}
\end{equation}
where $(x_1,\ y_1)$ and $(x_2,\ y_2)$ are the coordinates of the top-left and bottom-right corners.

\section{Methodology}
\label{sec:method}
To fully leverage multimodal evidence and ensure cross-modal consistency, we propose \textbf{M}odality-aware \textbf{C}onsistency \textbf{R}easoning (\textbf{MCR}) including \textbf{M}ulti-style \textbf{R}easoning \textbf{S}chema \textbf{I}njection (\textbf{MRSI}) and \textbf{C}onstraint-guided \textbf{V}erifiable \textbf{O}ptimization (\textbf{CVO}). The framework of MCR is illustrated in Figure~\ref{fig:method}. MRSI organizes the diverse reasoning schema with modality-specific constraints and enforces explicit reasoning, while CVO leverages the reasoning schema together with GRPO to further strengthen the model’s reasoning capability.

\subsection{Multi-style Reasoning Schema Injection}
To address the modality bias caused by insufficient cross-modal consistency reasoning, we propose MRSI, which injects constraint-centered and diverse reasoning schema into the inference process~\cite{expo,rest} to strengthen cross-modal verification. Specifically, MRSI is guided by four core constraints, covering entity recognition $\mathcal{C}_{s}$, type classification $\mathcal{C}_{t}$, visual entailment $\mathcal{C}_{e}$ and visual grounding $\mathcal{C}_{u}$:
\begin{equation}
\mathcal{C}=\{ \mathcal{C}_{s}, \mathcal{C}_{t}, \mathcal{C}_{e}, \mathcal{C}_{u}\}.
\label{eq:constraint}
\end{equation}
Each constraint aligns with the task and its relevant modality. See Appendix~\ref{sec:constraints_prompt} for an example. The resulting reasoning schema in Figure~\ref{fig:method} reflects both task- and modality- specific considerations.
As shown in Appendix~\ref{sec:reasoning_schema}, through templates, LLMs, or MLLMs, we transform $(s, v,\mathcal{C},\mathcal{Y}_{\tau})$ into programmatic reasoning steps $z$ with multiple styles on the labeled set $\mathcal{D}_{\mathcal{G}}$:
\begin{equation}
\mathcal{D}_{\mathcal{R}}=\bigcup_{(s,v,\mathcal{Y}) \in \mathcal{D}_{\mathcal{G}}}\Gamma_{\phi}\!\left(z \mid s,\ v, \ \mathcal{C}, \ \mathcal{Y}\right),
\label{eq:cot_dataset}
\end{equation}
where $\Gamma_{\phi}$ denotes template extractors, LLMs or MLLMs, and $\mathcal{D}_{\mathcal{R}}$ is the obtained CoT training dataset. The diversity of reasoning schema prevents the sampled trajectories from collapsing into overly similar outputs, avoiding the negligible advantages and gradient vanishing~\cite{collapse,yao2025r1}. We use $\mathcal{D}_{1}$ (a subset of $\mathcal{D}_{\mathcal{R}}$) to inject reasoning schema into MLLMs~~\cite{tang2025thinking,koksal2025milchat} via:
\begin{equation}
\begin{aligned}
\mathcal{L}_{\text{MRSI}}
= 
&-\,\mathbb{E}_{(x,v,z,y)\sim \mathcal{D}_1}\big[
\log \pi_{\text{MLLM}}(z \mid x,v)
\\
&{}+ \log \pi_{\text{MLLM}}(y \mid x,v,z)
\big],
\label{eq:cot_sft}
\end{aligned} 
\end{equation}
where $\pi_{\text{MLLM}}(z \mid x,v)$ and $\pi_{\text{MLLM}}(y \mid x,v,z)$ respectively denote the probability that MLLMs generate a reasoning path given the image–text pair and predict the answer based on the generated path. Through explicitly introducing $\mathcal{C}$ and $z$, MRSI compels the model to retain a reasoning path for cross-modal consistency checking.
\subsection{Constraint-guided Verifiable Optimization}
Reinforcement Learning with Verifiable Rewards~\cite{deepseek} replaces reward models with reward functions and has shown strong performance on reasoning tasks. Following this, we introduce CVO to  enhance cross-modal reasoning.
\subsubsection{Constraint-guided Verifiable Reward}
\label{sec:reward}
 Inspired by prior similar reward functions~\cite{liu2025visual,roit2023factually}, we anchor on $\mathcal{C}$ and design rule-based verifiable reward functions for GMNER, including entity count, entity span, entity type, entailment, and localization rewards.

\noindent\textbf{Entity Count, Span and Type Rewards.} The entity count reward $R_c$ encourages broader exploration while still maintaining precision, preventing the model from becoming overly conservative. It assigns a score by comparing the number of predicted entity triples with the ground-truth count. See  Appendix~\ref{sec:entity_count} for details. 

In computing the entity span reward $R_s$, we compute the token-level F1 score for every predicted–gold entity pair in a sample and perform optimal matching using the Hungarian algorithm. See Appendix~\ref{sec:token_level_f1} for details. The entity span reward for a sample is defined as the average token-level F1 over all matched pairs:
\begin{equation}
\begin{aligned}
R_s = \frac{1}{k} \sum_{(i,j)\in \mathcal{N}} F_{ij},
\label{eq:token_f1}
\end{aligned} 
\end{equation}
where $F_{ij}$ denotes the token-level F1 score between the $i$-th predicted entity and the $j$-th gold entity, $\mathcal{N}$ denotes the set of successfully matched entity pairs, and $k = |\mathcal{N}|$ is the number of matched pairs. And the type reward $R_\text{type}$ is formed as:
\begin{equation}
\begin{aligned}
R_t=\frac{1}{k}\sum_{i}^{k}\mathbbm{1} \{\hat t_i=t_i\},
\label{eq:type}
\end{aligned}
\end{equation}
where $\mathbbm{1}$ denotes the indicator function, which equals $1$ if the predicted type $\hat{t_i}$ matches the gold type $t_i$ and $0$ otherwise. The sample-level reward $R_t$ is the average over the $k$ matched pairs.

\noindent\textbf{Visual Grounding and Entailment Rewards.} The grounding reward function $R_u$ considers the Intersection-over-Union ($\text{IoU}$) metric, which measures the overlap between a predicted bbox and a gold bbox as the intersection area divided by their union. $R_u$ is formed as:
\begin{equation}
\begin{aligned}
R_u=\frac{1}{k}\sum_{i}^{k}\max(0,\ \frac{\text{IoU}_i - \sigma}{1-\sigma}),
\label{eq:iou}
\end{aligned}
\end{equation}
where $\text{IoU}_i$ denotes the $\text{IoU}$ between the $i$-th predicted and gold bbox, and we apply a threshold 
$\sigma$ and bounding boxes with an IoU below the threshold $\sigma$ are set 0 of IoU~\cite{liu2025visual}, while bounding boxes with an IoU above the threshold $\sigma$ are linearly mapped into $[0,1]$. The sample-level reward $R_u$ is the average over the $k$ matched pairs. And the entailment reward $R_e$ is formed as:
\begin{equation}
\begin{aligned}
& R_{e}= \frac{1}{k}\sum_{i=1}^{k} \mathbbm{1}\{\hat{v}_i = v_i\}, \\
& v_i= \mathbbm{1}\{l_i \neq \text{None}\},\ \  \hat{v}_i = \mathbbm{1}\{\hat{l}_i \neq \text{None}\},
\end{aligned}
\end{equation}
where $\mathbbm{1}$ denotes the indicator function, which returns $1$ if the condition inside the braces is true and $0$ otherwise, $v_i$ and $\hat{v}_i$ respectively indicate whether the gold and predicted entities are visible. For each matched entity, the reward is set to 1 if the predicted and gold locations are both None or both non-None; otherwise, it is 0. The sample-level reward $R_e$ is the average over the $k$ matched pairs.

Finally, our overall reward is a weighted combination of the above rewards:
\begin{equation}
R = \lambda_1R_c+\lambda_2R_s+\lambda_3R_t+\lambda_4R_u+\lambda_5R_e,
\label{eq:reward_total}
\end{equation}
where $\lambda_{j}$ denotes the weight of each reward term.
\subsubsection{Optimization with Verifiable Rewards}
Given verifiable rewards $R$ and remaining data $\mathcal{D}_2 = \mathcal{D}_{\mathcal{R}} \backslash \mathcal{D}_{1}$, CVO optimizes the policy to align cross-modal verification with the core constraints. For each query $q$ sampled, the current policy $\pi_{\theta_{\text{old}}}$ generates $G$ diverse responses $\{ o_1, o_2, \ldots, o_G \}$. The verifiable reward functions compute scores $\{ r_1, r_2, \ldots, r_G \}$ for each response by $R$. We then obtain the group advantage $A_i$:
\begin{equation}
\label{eq:advantage}
A_i = \frac{r_i - \mu_G}{\sigma_G},
\end{equation}
where $\mu_G$ denotes the empirical mean of the group rewards computed over $\{ r_1, r_2, \ldots, r_G \}$, $\sigma_G$ denotes their empirical standard deviation. To further improve training efficiency and reduce collapse risk, we apply sampling-based filtering to $\mathcal{D}_2$ based on reward distribution statistics. See the Appendix~\ref{sec:data_preparation} for details. CVO updates the policy with a GRPO style objective. The learning objective uses a clipped importance ratio to prevent overly aggressive updates and a length normalization to keep responses comparable. The objective function is defined as follows:
\begin{equation}
\begin{aligned}
&\mathcal{J}_{\text{CVO}}(\theta) 
= \mathbb{E}\!\left[q \sim P(Q),\; o \sim \pi_{\theta_{\text{old}}}(O \mid q)\right] \\
&\quad\frac{1}{|o|} \sum_{t=1}^{|o|}
   \min\!\Bigg(
      \frac{\pi_{\theta}(o_t \mid q, o_{<t})}{\pi_{\theta_{\text{old}}}(o_t \mid q, o_{<t})} A_t,
      \\
&\quad\operatorname{clip}\!\left(
          \frac{\pi_{\theta}(o_t \mid q, o_{<t})}{\pi_{\theta_{\text{old}}}(o_t \mid q, o_{<t})},
          1-\varepsilon,\,
          1+\varepsilon
      \right) A_t
   \Bigg),
\label{eq:loss}
\end{aligned}
\end{equation}
where clipping with threshold $\varepsilon$ prevents overly aggressive updates and length normalization ensures comparability across responses. The design yields stable group preference optimization without a critic and supports constraint anchored reasoning in multimodal settings.

\begin{table*}[ht]
\centering
\small
\setlength{\tabcolsep}{4pt}
\renewcommand{\arraystretch}{1.15}
\begin{tabular}{l|l ccc ccc ccc}
\toprule
\multirow{2}{*}{\textbf{Type}}  &
\multirow{2}{*}{\textbf{Methods}}  &
\multicolumn{3}{c}{\textbf{GMNER}} &
\multicolumn{3}{c}{\textbf{MNER}} &
\multicolumn{3}{c}{\textbf{EEG}} \\
\cmidrule(lr){3-5}\cmidrule(lr){6-8}\cmidrule(lr){9-11}
& & \textbf{Pre} & \textbf{Rec} & \textbf{F1}
& \textbf{Pre} & \textbf{Rec} & \textbf{F1}
& \textbf{Pre} & \textbf{Rec} & \textbf{F1} \\    
\midrule
\multirow{4}{*}{\textbf{Pipeline}} &
ITA\text{-}VinVL\text{-}EVG~\cite{wang2022ita}      & 52.4 & 50.8 & 51.6 & 80.4 & 78.4 & 79.4 & 56.6 & 54.8 & 55.7 \\
& BARTMNER\text{-}VinVL\text{-}EVG~\cite{gmner}  & 52.5 & 52.4 & 52.5 & 80.7 & 80.1 & 80.4 & 55.7 & 55.6 & 55.7 \\
& SCANNER~\cite{scanner}     & 68.3 & 68.7 & 68.5 &  -    &  -    &  -    &  -    &  -    &  -    \\
& ReFineG~\cite{tang2025refineg}   & 54.1 & 60.2 & 57.0 &      &  -    &  -    &  -    &  -    &  -    \\
& UnCo~\cite{tang2025unco}   & - & - & 64.6 &  -   &  -    &  81.7    &  -    &  -    &  69.6    \\

\midrule
\multirow{4}{*}{\textbf{Unified}} &
MNER\text{-}QG~\cite{mner_qg}        & 53.0 & 54.8 & 53.9 & 78.2 & 78.6 & 78.4 & 58.5 & 56.6 & 57.5 \\
& H\text{-}Index~\cite{gmner}        & 56.2 & 56.7 & 56.4 & 79.4 & 80.1 & 79.7 & 60.9 & 61.5 & 61.2 \\
& TIGER~\cite{finegrained}          & 55.8 & 57.5 & 56.6 & 79.9 & 80.1 & 80.3 & 60.7 & 61.8 & 61.3 \\
& MQSPN~\cite{multigrained}   & 59.0 & 58.5 & 58.8 & 81.2 & 79.7 & 80.4 & 61.9 & 62.9 & 62.4 \\
\midrule
\multirow{18}{*}{\textbf{End-to-End}} &
{\textbf{GLM4.5VL}~\cite{hong2025glm}}           & 33.0     & 44.4     & 37.8     & 43.0     & 57.9     & 49.4     & 36.2     & 48.7     & 41.6     \\
&\quad +CoT                         & 40.9 & 50.3 & 45.1 & 53.7 & 66.1 & 59.3 & 44.6 & 54.8 & 49.2 \\
& \quad +CoT+3\text{-}Shot           & 43.2 & 55.5 & 48.5 & 53.1 & 68.3 & 59.7 & 47.2 & 60.7 & 53.1 \\
& {\textbf{Qwen2.5VL\text{-}72B}~\cite{qwen2_5vl}}        & 24.0     & 44.5    & 31.2     & 32.2     & 59.8     & 41.9     & 26.3     & 48.9     & 34.2     \\
&\quad {+CoT}                         & 30.4 & 45.2 & 36.3 & 40.5 & 60.3 & 48.4 & 33.7 & 50.2 & 40.3 \\
& {\quad +CoT+3\text{-}Shot}            & 33.0 & 52.3 & 40.5 & 47.0 & 74.4 & 57.6 & 37.2 & 58.8 & 45.6 \\
\cdashline{2-11}
& {\textbf{Qwen2.5VL\text{-}7B}~\cite{qwen2_5vl}}        & 5.40     & 14.1     & 7.80     & 9.80     & 25.3     & 14.1     & 6.10     & 15.9    & 8.80    \\
&\quad {+CoT}                          & 11.4 & 13.6 & 12.4 & 20.2 & 24.0 & 21.9 & 12.9 & 15.4 & 14.1 \\
&\quad {+CoT+3\text{-}Shot}            & 16.5 & 33.7 & 22.2 & 27.5 & 56.2 & 37.0 & 18.3 & 37.3 & 24.5 \\
&\quad {+SFT}                          & 63.3 & 62.0 & 62.7 & 83.0 & 81.3 & 82.2 & 65.8 & 64.4 & 65.1 \\
&\quad\cellcolor{modelmint}{{+MRSI\,(\textbf{ours})}}                  & \cellcolor{modelmint}{69.1} & \cellcolor{modelmint}{68.1} & \cellcolor{modelmint}{68.6} & \cellcolor{modelmint}{82.4} & \cellcolor{modelmint}{81.2} & \cellcolor{modelmint}{81.8} & \cellcolor{modelmint}{72.1} & \cellcolor{modelmint}{71.0} & \cellcolor{modelmint}{71.5} \\
& \quad\cellcolor{modelmint}{+MRSI{+}CVO\,\textbf{(ours)}}           & \cellcolor{modelmint}\textbf{70.5} & \cellcolor{modelmint}\textbf{70.8} & \cellcolor{modelmint}\textbf{70.6} & \cellcolor{modelmint}\textbf{82.6} & \cellcolor{modelmint}\textbf{82.9} & \cellcolor{modelmint}\textbf{82.8} & \cellcolor{modelmint}\textbf{73.2} & \cellcolor{modelmint}\textbf{73.5} & \cellcolor{modelmint}\textbf{73.4} \\
& {\textbf{MimoVL\text{-}7B}~\cite{mimo}}             & 9.60     & 10.9     & 10.2     & 20.5     & 23.3     & 21.8     & 10.6     & 12.0     & 11.2     \\
&\quad {+CoT}             & 11.9     & 17.3     & 14.1     & 22.1     & 32.2     & 26.2     & 13.1     & 19.0     & 15.5     \\
&\quad {+CoT+3\text{-}Shot}             & 15.0     & 21.4     & 17.7     & 29.6     & 42.1     & 34.8     & 17.8     & 25.3    & 20.9     \\
&\quad {+SFT}                          & 63.5 & 60.6 & 62.0 & 81.7 & 78.0 & 79.8 & 67.0 & 63.9 & 65.4 \\
&\quad\cellcolor{modelcyan}{{+MRSI\,(\textbf{ours})}}                  & \cellcolor{modelcyan}{66.1} & \cellcolor{modelcyan}{65.5} & \cellcolor{modelcyan}{65.8} & \cellcolor{modelcyan}{81.5} & \cellcolor{modelcyan}{80.8} & \cellcolor{modelcyan}{81.1} & \cellcolor{modelcyan}{69.8} & \cellcolor{modelcyan}{69.2} & \cellcolor{modelcyan}{69.5} \\
& \quad\cellcolor{modelcyan}{+MRSI{+}CVO\,\textbf{(ours)}}           & \cellcolor{modelcyan}\textbf{69.4} & \cellcolor{modelcyan}\textbf{69.7} & \cellcolor{modelcyan}\textbf{69.6} & \cellcolor{modelcyan}\textbf{82.2} & \cellcolor{modelcyan}\textbf{82.5} & \cellcolor{modelcyan}\textbf{82.3} & \cellcolor{modelcyan}\textbf{72.8} & \cellcolor{modelcyan}\textbf{73.1} & \cellcolor{modelcyan}\textbf{72.9} \\
\bottomrule
\end{tabular}
\caption{\textbf{Comparison on GMNER, MNER and EEG.} \textbf{Pre}, \textbf{Rec} and \textbf{F1} respectively denote \textbf{Precision}, \textbf{Recall} and F1 score. Best in each block is bold.
 } 
\label{tab:gmner_mner_eeg}
\end{table*}

\section{Experiments}
We briefly introduce the datasets, baselines, and evaluation metrics used in our experiments, with further details provided in the Appendix~\ref{sec:more_experiemnt_detail}.
\label{sec:experiments} 
\subsection{Experimental Setup}
\textbf{Datasets.} Training datasets are from three sources: Twitter-GMNER~\cite{gmner} for GMNER, a multi-image MNER dataset MNER-MI~\cite{mner_mi}, and a visual grounding dataset, GREC~\cite{grec}. We evaluate Twitter-GMNER along with its two subtasks (MNER and EEG) in the main experiments, and conduct additional evaluations on MNER-MI and GREC. 

\noindent\textbf{Baselines.} Following prior work~\cite{multigrained}, we categorize existing approaches into pipeline and unified methods based on whether textual entity extraction and visual region prediction are executed within a single pass.
Furthermore, we investigate the applicability of open-source and close-source MLLMs within an end-to-end paradigm. 
To establish robust benchmarks, we implement \textit{Chain-of-Thought (CoT)}~\cite{cot}, \textit{Few-shot prompting}~\cite{brown2020language}, and \textit{Supervised Fine-tuning (SFT)}~\cite{ouyang2022training} as strong baselines for comparison.

\noindent\textbf{Evaluation Metrics.} Following~\citealp{gmner}, we evaluate GMNER, MNER and VG using Precision, Recall and F1 score. For the subtasks in GMNER, MNER identifies and classifies entities, while EEG grounds named entities in the image.

\subsection{Main Results}
As presented in Table~\ref{tab:gmner_mner_eeg}, regarding training-free approaches, the introduction of explicit reasoning via Chain-of-Thought (CoT) or Few-shot prompting yields notable performance gains compared to the direct application of MLLMs. Upon integrating our proposed MCR framework (with MRSI and CVO), all MLLMs consistently outperform existing baselines.
Specifically, on GMNER, the proposed method improves over the previous best unified method MQSPN~\cite{multigrained} by $11.87\%$ F1 scores and over the best pipeline method SCANNER~\cite{scanner} by $2.11\%$ F1 scores. 
Moreover, using Qwen2.5VL-7B, the proposed method respectively outperforms Qwen2.5VL-72B and. And the proposed method respectively surpasses direct Supervised Fine-Tuning (SFT) by $8.05\%$ and $7.57\%$ F1 scores on Qwen2.5VL-7B and MimoVL-7B. On MNER, the proposed method outperforms all methods at least $2.33\%$ F1 scores. On EEG, the proposed method exceeds the best unified method MQSPN by $10.97\%$ F1 scores.

\subsection{Performance on uni-modal bias dataset}
\label{sec:vg_mner}
Pipeline methods often decompose GMNER into MNER and VG, where the bidirectional modality biases in GMNER also manifest separately. Beyond GMNER, MCR further leverages datasets from these tasks for training and evaluation, using MNER-MI~\cite{mner_mi} for MNER and GREC~\cite{grec} for VG MNER-MI features weak text–image correlation, making visual bias more likely, while GREC may induce textual bias when a description corresponds to zero region. We evaluate SFT and MCR on MimoVL-7B and Qwen2.5VL-7B across these datasets.

\paragraph{Performance on MNER.} 
Table~\ref{tab:mner_mi_grec} show that MCR outperforms SFT on both tasks. Within MCR, the second-stage CVO generally surpasses the first-stage MRSI. This indicates that proposed method effectively leverages visual information to support entity extraction and classification while reducing the impact of irrelevant noise in the image.
\paragraph{Performance on VG.} 
In GREC, N-acc and Precision~\cite{grec} respectively evaluate grounding accuracy for cases with zero target region and with a target region. N-acc reflects the models' ability to judge entailment between text and image and thus partially characterizes textual bias. As shown in Table~\ref{tab:mner_mi_grec}, N-acc improves across models, indicating that our method effectively reduces text bias in MLLMs.

\begin{table}[t]
    \centering
    \small
    \setlength{\tabcolsep}{5pt}
    \renewcommand{\arraystretch}{1.15}
    \resizebox{\columnwidth}{!}{%
    \begin{tabular}{lccc cccc}
    \toprule
      \multirow{2}{*}{\textbf{Methods}}
      & \multicolumn{3}{c}{\textbf{MNER-MI}}
      & \multicolumn{2}{c}{\textbf{GREC-testA}}
      & \multicolumn{2}{c}{\textbf{GREC-testB}} \\
    \cmidrule(lr){2-4} \cmidrule(lr){5-6} \cmidrule(lr){7-8}
      & \textbf{Pre} & \textbf{Rec} & \textbf{F1}
      & \textbf{N-acc} & \textbf{Pre}
      & \textbf{N-acc} & \textbf{Pre} \\
    \midrule

    \multicolumn{1}{l}{\textbf{Qwen2.5VL-7B}}  & - & - & - & - & - & - & - \\ 
    \quad+SFT        
      & 84.0 & 83.6 & 83.8
      & 70.6 & 81.1
      & 70.2 & 62.5 \\
    \cellcolor{modelmint}\quad+MRSI        
      & \cellcolor{modelmint}82.1 & \cellcolor{modelmint}81.8 & \cellcolor{modelmint}82.0
      & \cellcolor{modelmint}74.2 & \cellcolor{modelmint}89.1
      & \cellcolor{modelmint}70.3 & \cellcolor{modelmint}71.5 \\
    \cellcolor{modelmint}\quad+MRSI{+}CVO 
      & \cellcolor{modelmint}84.1 & \cellcolor{modelmint}\textbf{86.2} & \cellcolor{modelmint}\textbf{85.1}
      & \cellcolor{modelmint}74.7 & \cellcolor{modelmint}\textbf{90.4}
      & \cellcolor{modelmint}69.7 & \cellcolor{modelmint}72.8 \\
    \midrule

    \multicolumn{1}{l}{\textbf{MimoVL-7B}}  & - & - & - & - & - & - & - \\
    \quad+SFT        
      & 81.8 & 80.8 & 81.23
      & 65.7 & 66.1
      & 69.6 & 45.6  \\
    \cellcolor{modelcyan}\quad+MRSI        
      & \cellcolor{modelcyan}83.2 & \cellcolor{modelcyan}83.4 & \cellcolor{modelcyan}83.3
      & \cellcolor{modelcyan}72.6 & \cellcolor{modelcyan}89.4
      & \cellcolor{modelcyan}68.6 & \cellcolor{modelcyan}71.6 \\
    \cellcolor{modelcyan}\quad+MRSI{+}CVO 
      & \cellcolor{modelcyan}\textbf{84.7} & \cellcolor{modelcyan}85.0 & \cellcolor{modelcyan}84.8
      & \cellcolor{modelcyan}\textbf{75.7} & \cellcolor{modelcyan}90.4
      & \cellcolor{modelcyan}\textbf{71.9} & \cellcolor{modelcyan}\textbf{73.0} \\
    \bottomrule
    \end{tabular}}
    \caption{\textbf{Results on MNER-MI and GREC.} For GREC dataset, we remove cases where a single textual description corresponds to multiple regions in the image.}
    \label{tab:mner_mi_grec}
\end{table}

\subsection{Component Analysis}
\begin{table}[h]   
\centering
\resizebox{\columnwidth}{!}{%
\begin{tabular}{l ccc ccc ccc}
\toprule
\multirow{2}{*}{\textbf{Methods}} &
\multicolumn{3}{c}{\textbf{GMNER}} &
\multicolumn{3}{c}{\textbf{MNER}} &
\multicolumn{3}{c}{\textbf{EEG}} \\
\cmidrule(lr){2-4}\cmidrule(lr){5-7}\cmidrule(lr){8-10}
& \textbf{Pre} & \textbf{Rec} & \textbf{F1}
& \textbf{Pre} & \textbf{Rec} & \textbf{F1}
& \textbf{Pre} & \textbf{Rec} & \textbf{F1} \\
\midrule
Ours                                & \textbf{70.5} & \textbf{70.8} & \textbf{70.6} & \textbf{82.6} & \textbf{82.9} & \textbf{82.8} & \textbf{73.2} & \textbf{73.5} & \textbf{73.4} \\
w/o MRSI                             & 50.2 & 53.2 & 51.7 & 64.3 & 68.1 & 66.2 & 54.2 & 57.5 & 55.8 \\
w/o CVO                             & 68.9 & 68.6 & 68.8 & 81.9 & 81.5 & 81.7 & 72.2 & 71.9 & 72.0 \\
w/o $\mathcal{D}_{\mathcal{R}}$     & 69.5 & 68.8 & 69.2 & 81.6 & 80.7 & 81.1 & 72.8 & 72.1 & 72.5 \\
w/o Inst                     & 67.3 & 66.1 & 66.7 & 81.7 & 80.2 & 80.9 & 70.8 & 69.5 & 70.1 \\
\bottomrule
\end{tabular}}
\caption{\textbf{Ablation results on GMNER, MNER, and EEG.} w/o $\mathcal{D}_{\mathcal{R}}$ means removing diverse reasoning styles and training MCR with a single style. w/o Inst means removing the components that specify stepwise cross-modal verification and cautionary guidelines from the original instructions.}
\label{tab:ablation}
\end{table}

\begin{table}[h]
\centering
\small
\setlength{\tabcolsep}{4pt}
\renewcommand{\arraystretch}{1.0}
\resizebox{0.9\columnwidth}{!}{%
\begin{tabular}{l l c c}
\toprule
\textbf{Model} & \textbf{Method} & \textbf{N-Rate (\%)} & \textbf{N-Count} \\
\midrule
\multirow{3}{*}{Qwen2.5VL-72B} 
& Direct Prompt & 29.2& 1372 \\
& CoT & 15.6 & 610 \\
& CoT+3-Shot & 5.8 & 241 \\
\midrule
\multirow{3}{*}{GLM4.5VL} 
& Direct Prompt & 26.9 & 898 \\
& CoT & 24.3 & 820 \\
& CoT+3-Shot & 13.8 & 464 \\
\midrule
\multirow{2}{*}{Qwen2.5VL-7B} 
& CoT+3-Shot & 13.2 & 363 \\
& MCR(\textbf{ours}) & 0.1 & 3 \\
\midrule
\multirow{2}{*}{MimoVL-7B} 
& CoT+3-Shot & 24.3 & 637 \\
& MCR(\textbf{ours}) & 0.2 & 5 \\
\bottomrule
\end{tabular}
}
\caption{\textbf{Quantitative results of visual bias.} \textbf{N-Count} means the number of recalled entities that are absent from the sentence and \textbf{N-Rate} means the proportion of such entities among all recalled entities. Direct Prompt denotes a concise task instruction.}
\label{tab:n_rate}
\end{table}

\paragraph{Compelling the model to reason is essential.} In Table~\ref{tab:ablation}, w/o MRSI directly applying CVO on $\mathcal{D}_{\mathcal{R}}$ while skipping MRSI, and we observe that the F1 score drops by $18.95\%$. This indicates the stepwise cross-modal verification path established by MRSI is a necessary prerequisite.

\paragraph{Verifiable rewards can further improve the model's reasoning capability.} As shown by w/o CVO in Table~\ref{tab:ablation}, using the CVO data to continue MRSI yields only a marginal $0.18\%$ improvement. In contrast, strengthening the model's reasoning with CVO adds to a $2.04\%$ gain. This contrast highlights the necessity of CVO for further enhancing the model’s cross-modal verification capability.

\paragraph{Multi-style reasoning schema enhance training performance.} w/o $\mathcal{D}_{\mathcal{R}}$ eliminates reasoning diversity, and training MCR with a single reasoning style. It slows down the CVO stage and results in smaller performance gains. This degradation is caused by reduced coverage and exploration of reasoning trajectories and shortcut reliance.

\paragraph{Reasoning-related instruction components effectively guide the model.} w/o Inst removes the components that specify stepwise cross-modal verification and cautionary guidelines for MCR training from the original instructions. Without this constraint-aware guidance, the model struggles to establish clear intermediate goals and consistent verification criteria, which weakens execution fidelity at key steps and results in a $3.90\%$ drop in F1 score under the same training conditions.

\subsection{Further Analysis}
\begin{figure}[t]
  \centering
   \includegraphics[width=1.0\linewidth]{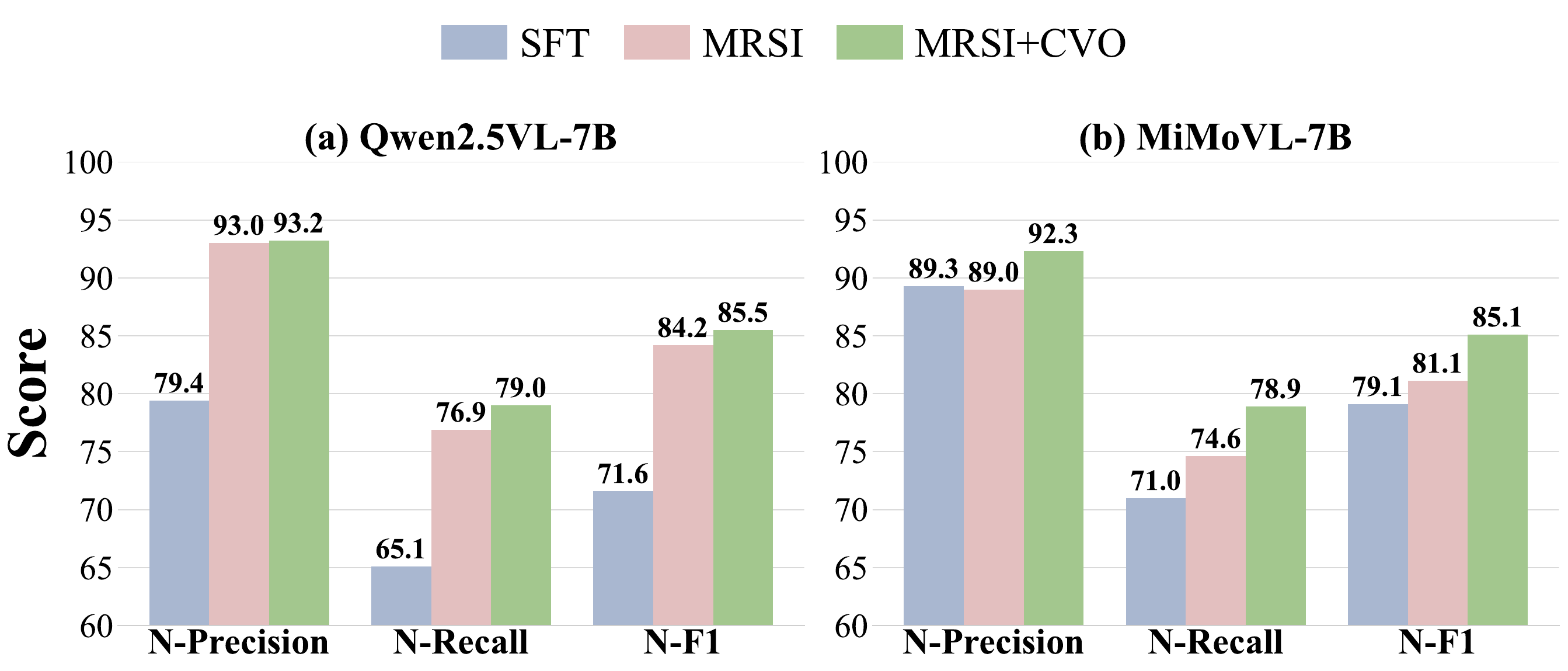}
   \caption{\textbf{Quantitative results of textual bias.} MCR effectively improves the models' ability to determine whether an entity is present, which in turn indicates that MCR mitigates textual bias.}
   \label{fig:text_bias}
\end{figure}
\begin{figure}[t]
  \centering
   \includegraphics[width=1.0\linewidth]{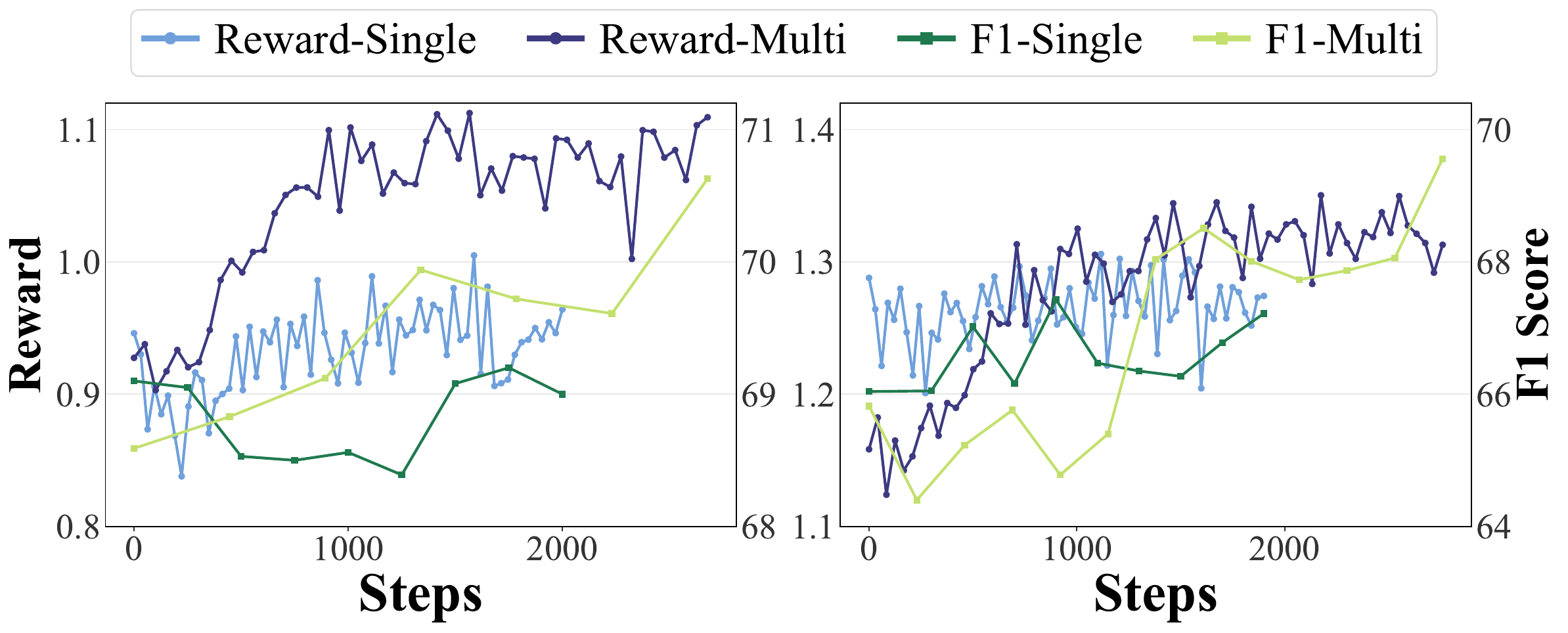}
   \caption{\textbf{Effect of Multi-style vs. Single-style Reasoning Schema on F1 and Reward Scores in CVO on Qwen2.5VL-7B (Left) and MimoVL-7B (Right).} Single means single-style reasoning schema and Multi means multi-style reasoning schema. }
   \label{fig:reward_f1}
\end{figure}
\begin{figure}[h]
  \centering
   \includegraphics[width=1.0\linewidth]{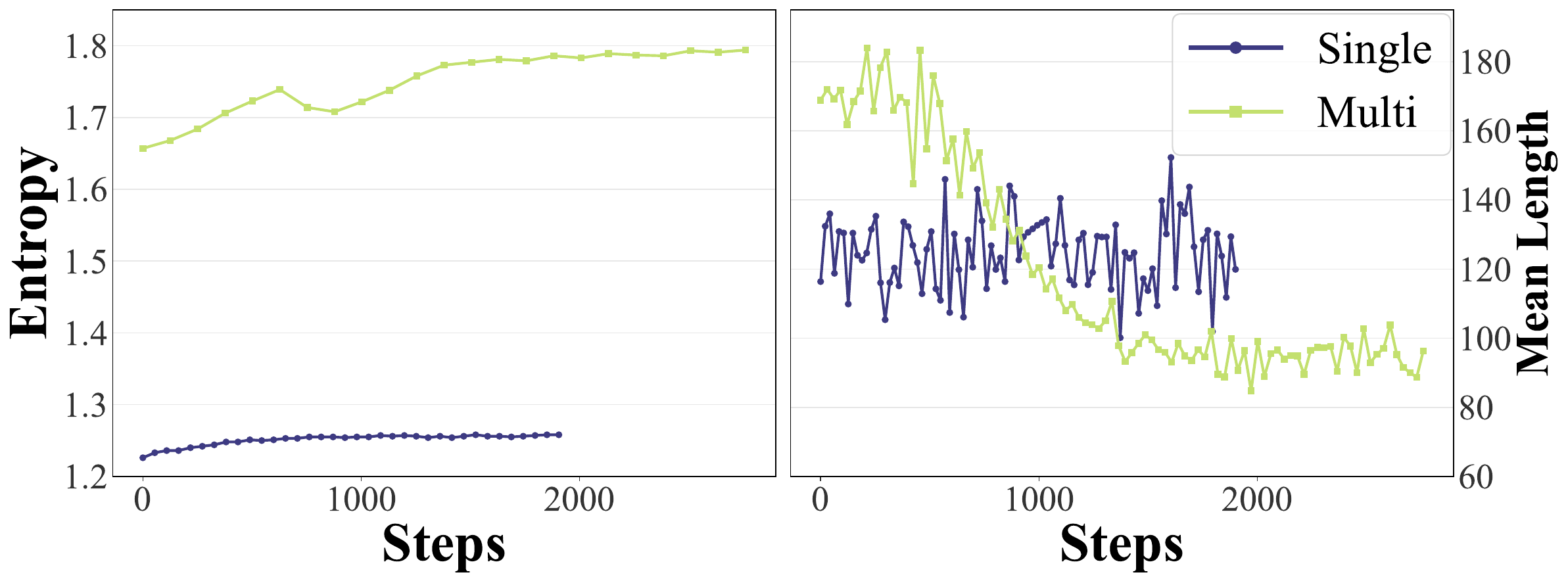}
   \caption{\textbf{Effect of Multi-style vs. Single-style Reasoning Schema on Cross Entropy (Left) and Mean Completion Length (Right) in CVO on MimoVL-7B.} }
   \label{fig:mimo_entropy_completion}
\end{figure}
\paragraph{MCR effectively mitigates visual bias in GMNER.} 
Directly inspecting every test image to quantify how MCR handles visual bias is impractical, so we introduce two indirect metrics. Based on whether a recalled entity appears in the input sentence, we define \textbf{N-Count} as the number of recalled entities that are absent from the sentence, and \textbf{N-Rate} as the proportion of such entities among all recalled entities. As shown in Table~\ref{tab:n_rate}, training-free approaches such as CoT and few-shot prompting can partially alleviate visual bias in GMNER. In contrast, MCR reduces visual bias for Qwen2.5VL-7B and MimoVL-7B to a near-negligible level.
\noindent\paragraph{MCR effectively mitigates textual bias in GMNER.}
Inspired by the N-acc metric in GREC, we introduce three metrics to quantify textual bias in GMNER. \textbf{N-Pre} measures the fraction of predicted text-only triples with location ``None'' that correctly match gold text-only triples, while \textbf{N-Rec} measures the proportion of gold text-only triples that the model correctly predicts as having no location. \textbf{N-F1} is the harmonic mean of N-Pre and N-Rec. As shown in Figure~\ref{fig:text_bias}, MCR on Qwen2.5VL-7B improves over SFT by nearly $+14\%$ across all three metrics, indicating effective mitigation of textual bias. Notably, N-Pre is substantially higher than N-Rec, suggesting that the model conservatively recalls text-only entities to achieve more accurate entailment judgments.

\noindent\paragraph{Multi-style reasoning schema help improve the training effectiveness of CVO.} Figure~\ref{fig:reward_f1} compares single-style and multi-style reasoning schema during CVO training in terms of both reward and F1 score. At the early stage, single-style reasoning achieves higher rewards and F1 scores due to more focused supervision from MRSI. However, we observe that multi-style reasoning yields higher rewards and F1 scores finally, indicating its stronger ability to stimulate exploration and support more effective policy optimization. Moreover, it leads to more stable optimization, while single-style reasoning suffers from larger fluctuations.

\noindent\paragraph{Controlled Policy Exploration in CVO.} As shown in Figure~\ref{fig:mimo_entropy_completion}, when multi-style reasoning schema is used in CVO, the cross-entropy increases gradually and then stabilizes, indicating that the model performs controlled and effective exploration over diverse reasoning strategies. Meanwhile, the mean completion length first decreases and then converges, suggesting that the model gradually settles into more concise and consistent reasoning patterns. In contrast, when CVO is trained with single-style reasoning schema, the cross-entropy exhibits only a very slow increase, while the mean completion length fluctuates without clear convergence.

\begin{figure}[t]
  \centering
   \includegraphics[width=1.0\linewidth]{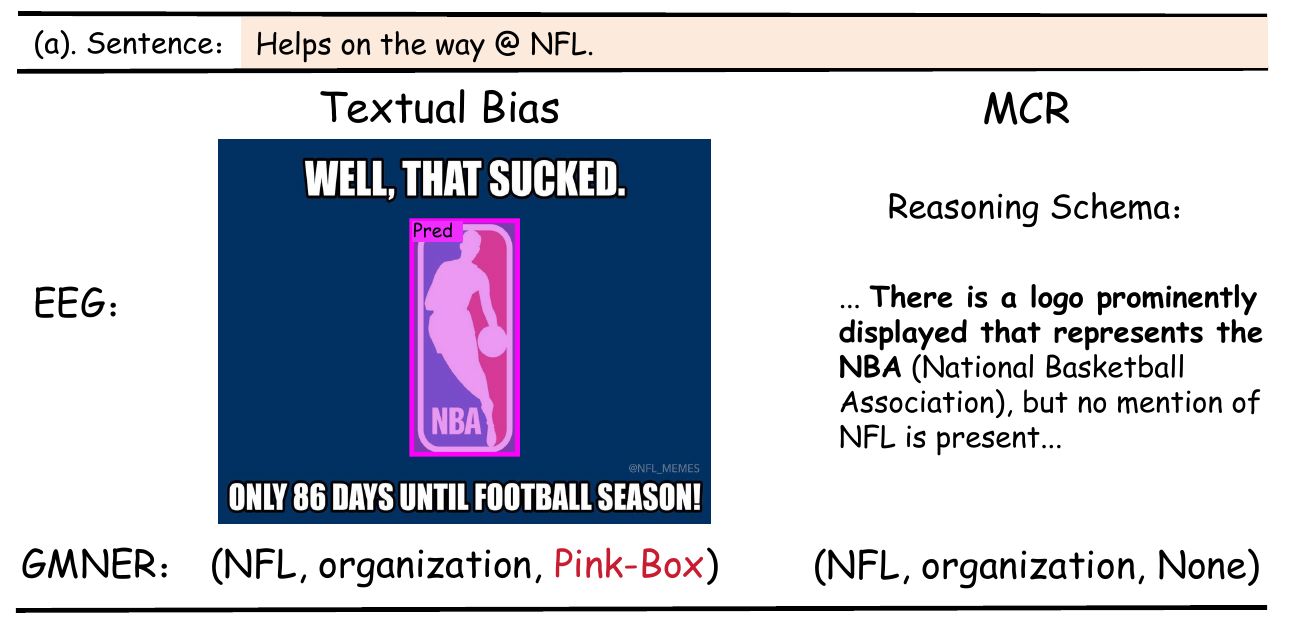}
   \caption{Case Studies of MCR Mitigating Modality Bias. 
   }
   \label{fig:case_1}
\end{figure}

\noindent\paragraph{Case Study.} 
As shown in Figure~\ref{fig:case_1}, Naive End-to-end methods may incorrect grounding due to insufficient cross-modal verification, such as grounding the “NBA” logo to the textual entity “NFL.” MCR mitigates this by explicitly reasoning over image–text consistency. More cases are provided in Appendix~\ref{sec:case_study}.
\section{Conclusion}
In this work, we advance GMNER by reformulating it as an end-to-end generative reasoning task.
We diagnose a critical pathology—\textit{modality bias}—revealing that MLLMs often rely on unimodal cognitive shortcuts rather than rigorous cross-modal verification. 
To address this, we propose Modality-aware Consistency Reasoning (\textbf{MCR}), which enforces structured cross-modal reasoning to mitigate modality bias.
Comprehensive evaluations on GMNER, MNER, and Visual Grounding benchmarks demonstrate that MCR effectively mitigates modality biases, enabling rigorous cross-modal verification and achieves superior performance compared to existing baselines.
Besides, a comprehensive suite of ablation experiments validates the necessity of our design choices and the stability of the optimization mechanism.

\section{Limitations}
Despite the promising performance of MCR in mitigating modality bias across GMNER, MNER, and VG tasks, our framework remains constrained by the inherent parametric knowledge limits of the underlying MLLMs. Specifically, MCR relies on the model's internal knowledge base for entity recognition; consequently, it may struggle to generalize to unseen entities that are absent from the pre-training corpus.


\bibliography{custom}
\appendix
\clearpage
\setcounter{page}{1}
\section{Ethical Considerations}

\subsection{Potential Risks}
Potential risks associated with our work include the misuse of entity grounding capabilities for surveillance purposes and the possibility of model hallucinations leading to misinformation. We mitigate these risks by using only publicly available datasets and strictly filtering harmful content, but we urge practitioners to exercise caution and respect user privacy when deploying these models in real-world applications.
\subsection{Use of LLM}
In the preparation of this manuscript, we utilized Large Language Models (LLMs) for grammatical error correction and polishing to improve readability.
\subsection{Code and Data}
All images and generated text contexts, which we use to train MCR, strictly follow guidelines designed to exclude any harmful, unethical, or offensive content. 
Furthermore, the data used in MCR does not involve any comparisons of harmful, ethical, or offensive content between image pairs.
Our code and curated dataset annotations will be released under an open-source license (e.g., MIT or CC-BY 4.0) upon acceptance, and we strictly adhere to the licensing terms and usage policies of the original datasets (Twitter-GMNER, MNER-MI, GREC) and backbone models used in this work.
\section{Modality-specific Constraints}
\label{sec:constraints_prompt}
Here is one of the instructions used for distillation, which requires the model to consider the relevant modalities during execution and to produce results consistent with the labels.
\begin{tcolorbox}[colback=gray!5!white,colframe=black!75!white,title=Modality- and Task- specific Constraints]
\small
Grounded Multimodal Named Entity Recognition (GMNER) task requires, given a text and a paired image, \textbf{recognize the meaningful and specific entities from the text} that are ... . \textbf{Entity type classification is primarily based on textual information}, with \textbf{visual cues considered when the text is insufficient to make a confident determination}. When you predict each entity's location, \textbf{if the entity appears in the image}, the corresponding location is a bounding box (bbox); \textbf{if the entity does not appear in the image}, the corresponding location is None ...\\
Given a piece of text and its paired image:<image><sentence>.\\ 
The ground-truth labels for this image-text pair in the GMNER task are: <label>.\\
Now \textbf{generate the reasoning process that leads from the image-text pair to the ground-truth labels}. The content of the thought process should be your reasoning on how to obtain the true label \textbf{based on the image and text input}.
\end{tcolorbox}

\section{Multiple Styles Reasoning Schema}

\label{sec:reasoning_schema}
We construct diverse reasoning styles and paths using templates, LLMs, and MLLMs, and we design corresponding prompts for each style. We next illustrate the procedure with the GMNER task.
\subsection{Instruction}
\label{sec:instruction}
To ensure the model understands the task, we first design a task-introduction instruction and use it across all experiments.
\begin{tcolorbox}[colback=gray!5!white,colframe=black!75!white,title=Instruction for GMNER, label=box:instruction
]
\small
Here is a Grounded Multimodal Named Entity Recognition task. Given a text and a paired image, You need to identify all entities in the given text, assign each entity a category, and locate the corresponding entities in the image during the entity prediction.
\end{tcolorbox}
\noindent\textbf{Instructions to format the thought process.} To ensure that the generated reasoning is produced in a formatted style constrained by fixed tags, we design two instruction prompts for distinct reasoning routes, and we next present one of the instructions.
\begin{tcolorbox}[colback=gray!5!white,colframe=black!75!white,title=Formal Type Prompt for GMNER, label={box:formal}
]
\small
To accomplish this task, follow the steps below and place your reasoning and the results of each step inside the <process></process> tags. \\
1.First, prioritize the textual information; use the image only as a supplement. From the text, identify how many entities are present (zero or more) and list them. Important: do NOT extract entities that appear only in the image but not in the text, and do NOT omit entities that appear in the text but not in the image. Put the number of entities inside \texttt{<entity\_num></entity\_num>} tags. \\
2.Second, determine the type of each extracted entity. Use one of these labels: person, organization, location, miscellaneous. Put each entity and its type inside a \texttt{<mner></mner>} tag in the format: (entity text, entity type). \\
3.Third, decide whether each entity is visible in the image. For entities not visible in the image, place (entity text, invisible) inside an \texttt{<entailment></entailment>} tag. For entities that are visible, place (entity text, visible) inside an \texttt{<entailment></entailment>} tag. \\
4.Fourth, provide location information: for visible entities, give their bounding box as (x1, y1, x2, y2); for invisible entities, use None. Put each item inside a \texttt{<location></location>} tag in the format:(entity text, (x1, y1, x2, y2)) for visible entities;(entity text, None) for invisible entities. \\
After completing the steps above, synthesize your findings and produce the final answer of the task inside \texttt{<answer></answer>} tags.
\end{tcolorbox}

\noindent\textbf{Instructions to output thought process from LLMs.}
To elicit reasoning in a question–answer or few-conclusion style, we design two instruction prompts for distinct reasoning routes, and we next present one of the instructions. The LLM-augmented reasoning paths are generated using similar prompts.
\begin{tcolorbox}[colback=gray!5!white,colframe=black!75!white,title=Conclusion Type Prompt for GMNER, label=box:conclusion
]
\small
To accomplish this task, you need to follow the reasoning rules below to carry out step-by-step reasoning and accomplish the task objectives. \\
Goal and Reasoning rules: \\
1. Extract all special entities that appear in the TEXT (do not invent entities that only appear in the image but not in the text). \\
2. Assign a type to each special entity from (person, organization, location, miscellaneous).
3. Decide if each special entity is visible in the IMAGE. \\
4. If visible, provide the bbox coordinates as (x1, y1, x2, y2) of the special entity in the IMAGE; if not visible, use None as location information. \\
Put your reasoning in <process></process> tags and put the final answer in format of ( entity text, entity type, None or bounding box ) in <answer></answer>.
\end{tcolorbox}

\noindent\textbf{Instructions to output distilled thought process.}
We use the follow instruction to make MLLMs output distilled style reasoning paths.
\begin{tcolorbox}[colback=gray!5!white,colframe=black!75!white,title=Distill Type Prompt for GMNER, label=box:distill
]
\small
To accomplish this task, you need to carefully analyze this task step by step from the perspectives of the meaning of special entities, text semantics, image content, whether the special entities appear in the image, and the corresponding location information, and place the thought process within <process></process> tags, and put the triplet results in format of ( entity text, entity type, None or bounding box ) within <answer></answer> tags.
\end{tcolorbox}

\subsection{Reasoning Styles and Paths}
\label{sec:styles}
The above diversified instructions yield varied reasoning styles and paths, and we next illustrate two example reasoning processes on a single sample. Different reasoning styles or paths produce different output formats, and even within the same sample, the ordering of entity triples as well as the ordering of the entity, type, and location within each triple can vary.
\begin{tcolorbox}[colback=gray!5!white,colframe=black!75!white,title=Thought Process v1
]
\small
<process>
Let's analyze both the text and the image step by step: \\
1. Identify Entities in the Text: The text mentions "Spurs" and "Premier League". \\
2. Determine Entity Types: "Spurs" refers to Tottenham Hotspur Football Club, which is an organization. "Premier League" is also an organization, specifically a professional football league. \\
3. Check Visibility in the Image: In the image, there is a visible logo and text related to "Spurs". The logo of a horse and the word "SPURS" can be clearly seen on the cake box. There is no visible mention or depiction of  "Premier League" in the image. \\
4. Locate Entities in the Image: For "Spurs", the logo and text are present. By examining the image, the bounding box coordinates for "Spurs" can be determined as (405, 216, 558, 324), which encompasses the area where the "Spurs" logo. Since "Premier League" is not visible in the image, its location is marked as None. \\
5. Compile the Labels: \\
(Spurs, organization, (405, 216, 558, 324)) \\
(Premier League, organization, None) \\
</process>

\end{tcolorbox}
\begin{tcolorbox}[colback=gray!5!white,colframe=black!75!white,title=Thought Process v2]
\small
<process> \\
    In the text, I've identified "Spurs" as an entity, which appears to be visible in the image at the coordinates (405, 216, 558, 324). Based on my understanding, "Spurs" refers to Tottenham Hotspur Football Club, an organization in the English football league system.
    I also recognize "Premier League" as an entity in the text, though it doesn't seem to be visually present in the image. The Premier League is the top level of the English football league system. \\
    In total, there are two entities in the text. \\
</process>
\end{tcolorbox}

\section{More Details about Rewards}
\subsection{Entity Count Rewards}
\label{sec:entity_count}
After MRSI, the model tends to conservatively recall entities to avoid visual bias. To increase recall while preventing visual bias from reappearing, we introduce a entity count reward defined by the difference between the predicted and gold entity counts. The penalty is scaled by the true count: harsher penalties are applied when the true count is small, while penalties are more lenient when the true count is large. The reward is formed as:
\begin{equation}
w_o(q) =
\begin{cases}
0.4, & 1 \le q \le 2,\\
0.2, & 3 \le q \le 4,\\
0.1, & q \ge 5,\\
0,   & \text{otherwise.}
\end{cases}
\end{equation}
\begin{equation}
w_u(p) =
\begin{cases}
0.5, & 0 \le p \le 2,\\
0.3, & 3 \le p \le 4,\\
0.2, & p \ge 5,\\
0,   & \text{otherwise.}
\end{cases}
\end{equation}
\begin{equation}
R_c =
\begin{cases}
1, & p = q,\\[2pt]
\max\bigl(0,\; 1 - (p-q) w_o(q)\bigr),  & p > q > 0,\\[2pt]
\max\bigl(0,\; 1 - (q-p) w_u(p)\bigr), & 0 < p < q, \\[2pt]
0, & \text{otherwise.}
\end{cases}
\label{eq:location_format}
\end{equation}
where $p$ and $q$ respectively denote the numbers of predicted and gold entities in a sample, $w_o$ and $w_u$ are the penalty weights for excessive recall and insufficient recall. The reward $R_\mathrm{count}$ is computed as a function of the relationship between $p$ and $q$.

\subsection{Token-level F1 score}
\label{sec:token_level_f1}
For each predicted entity span $\hat{e}_{i}=\{\hat{e}_{i,1},\ldots,\hat{e}_{i,n}\}$ and gold entity span $e_{j}=\{e_{j,1},\ldots,e_{j,m}\}$, where $\hat{e}_{i,k}$ and $e_{j,k}$ denote the $k$-th token in the predicted and gold spans respectively, we first compute the length of their longest contiguous token overlap $w_{ij}$ between them. Then, we define the token-level precision and recall as:
\begin{equation}
P_{ij}=\frac{w_{ij}}{n}, \ R_{ij}=\frac{w_{ij}}{m}
\label{eq:token_pre_rec}
\end{equation}
where $n$ and $m$ respectively denote the numbers of tokens in the predicted and gold spans. The token-level F1 score for this pair is finally computed as:
\begin{equation}
F_{ij}=\frac{2P_{ij}R_{ij}}{P_{ij} + R_{ij}}.
\label{eq:token_f1}
\end{equation}
Given the token-level F1 matrix $\{F_{ij}\}$ over all predicted–gold span pairs, we apply the Hungarian algorithm to obtain an optimal one-to-one matching between predicted and gold entities. Let $\mathcal{M}$ denote the set of matched index pairs $(i,j)$ and $k = |\mathcal{N}|$ be the number of matched pairs. The entity span reward for a sample is then defined as the average token-level F1 over all matched pairs:
\begin{equation}
\begin{aligned}
R_s = \frac{1}{k} \sum_{(i,j)\in \mathcal{N}} F_{ij}.
\label{eq:token_f1}
\end{aligned} 
\end{equation}

\subsection{Data Preparation}
\label{sec:data_preparation}
To prevent over-reliance on fixed templates and mitigate training collapse~\cite{collapse} in CVO, we construct a multi-style set of reasoning schema $\mathcal{D}_{\mathcal{R}}$ during MRSI. We then use only a subset $\mathcal{D}_1$ for MRSI, and allocate the remainder $\mathcal{D}_2 = \mathcal{D}_{\mathcal{R}} \backslash \mathcal{D}_{1}$ to CVO for calibrating and optimizing cross-modal verification on core constraints. To further improve training efficiency and reduce collapse risk, we apply sampling-based filtering to $\mathcal{D}_2$. For each sample’s $G$ responses $\{ o_1, o_2, \ldots, o_G \}$ with rewards $\{ r_1, r_2, \ldots, r_G \}$, we compute the standard deviation, maximum reward, and median reward, and impose preset thresholds on these statistics. Specifically, only samples with a standard deviation of at least 0.1, a maximum group reward of at least 0.8, and a median group reward between 0.08 and 0.6 are retained for training.~\cite{rest}. 

\section{More Experiment Details}
\label{sec:more_experiemnt_detail}

\subsection{Datasets}
\label{sec:datasets_detail}
Our training data are drawn from three sources: Twitter-GMNER~\cite{gmner} for GMNER and NER, a multi-image multimodal NER dataset MNER-MI~\cite{mner_mi}, and generalized visual grounding dataset GREC~\cite{grec}. Because GREC includes cases where one textual description corresponds to multiple image regions, which is incompatible with the GMNER setting, we exclude samples in which a single textual description corresponds to multiple image regions. We evaluate Twitter-GMNER in the main experiments, and conduct additional evaluations on MNER-MI and GREC in Section~\ref{sec:vg_mner}. As shown in Table~\ref{tab:dataset_stats}, we also report the number of raw datasets used during training and evaluation. For GMNER and MNER-MI, we use the full datasets. For GREC, we first filter out multi-target cases, then select 14,000 samples from the remaining data for training, while retaining all remaining validation and test samples. In total, we use 55,712 samples annotated with multi-style reasoning schema for training.
\begin{table}[t]
\centering
\begin{tabular}{lccc}
\toprule
Dataset & Train & Val & Test \\
\midrule
GMNER   & 7000  & 1500 & 1500 \\
MNER-MI & 6856  & 860  & 860  \\
GREC    & 14000 & 5309 & 19066 \\
\bottomrule
\end{tabular}
\caption{Dataset statistics used in our experiments.}
\label{tab:dataset_stats}
\end{table}
\subsection{Baselines}
\label{sec:baselines_detail}
Following prior work~\cite{multigrained}, we categorize GMNER approaches into unified and pipeline methods. Unified methods use pretrained language models to extract entity–type–location triples in a single pass, while pipeline methods decompose the process into multiple stages handled by different models. Unified approaches reduce error propagation and improve over early pipelines~\cite{wang2022ita,gmner}, but recent pipeline methods that incorporate LLMs as knowledge bases achieve substantially better performance~\cite{llms_as_bridges,scanner}. In contrast to prior unified methods that still rely on auxiliary components and employ LLMs as auxiliary tools, we propose an end-to-end unified approach that uses MLLMs to complete all steps in a single inference.
\paragraph{Pipeline Methods.} (1) \textbf{ITA-VinVL-EVG}~\cite{wang2022ita} formulates multimodal named entity recognition as an image–text alignment problem. (2) \textbf{BARTMNER-VinVL-EVG}~\cite{gmner} first uses generative model BART to identify entity type pairs, and then uses the Entity Extraction \& Grounding (EEG) model to predict the bounding box for each pair. (3) \textbf{Scanner}~\cite{scanner} first identifies textual and visual entities using NER and visual grounding models, enriches entity semantics with LLMs and external knowledge bases, and finally matches textual entities to visual locations via a trained module. (4) \textbf{UnCo}~\cite{tang2025unco} adopts an uncertainty-aware collaboration between small models and large multimodal language models to refine grounded multimodal named entity recognition predictions. (5) \textbf{ReFineG}~\cite{tang2025refineg} combines small supervised models and large language models to enhance low-resource grounded multimodal named entity recognition through refinement and knowledge transfer.
\paragraph{Unified Methods.} (1) \textbf{MNER-QG}~\cite{mner_qg} formulates multimodal named entity recognition as an unified machine reading comprehension task, where entity queries are grounded to both textual context and visual evidence. (2) \textbf{H-index}~\cite{gmner} formulates GMNER to a sequence generation task with a multimodal BART model. (3) \textbf{TIGER}~\cite{finegrained} formulates fine-grained named entity recognition and grounding as a sequence generation task by converting entity-type-object triples into target text and employing T5 model to jointly predict entity spans, fine-grained types, and corresponding image objects. (4) MQSPN~\cite{multigrained} formulates grounded multimodal named entity recognition as a set prediction problem that employs multi-grained learnable queries to explicitly align textual entities with visual regions.
\paragraph{End-to-end Methods.} 
GLM4.5VL, Qwen2.5VL, and MimoVL are multimodal large language models with strong capabilities in multimodal understanding, reasoning, and visual grounding. We evaluate these models under different prompting and training settings, including direct instruction prompting, Chain-of-Thought (CoT) prompting, and CoT with 3-shot demonstrations. In addition, we include a supervised fine-tuning (SFT) baseline, where the models are fine-tuned on GMNER training data.

\subsection{Evaluation Metrics}
\label{sec:metric_detail}
\paragraph{GMNER.}
For GMNER and its two subtasks, MNER and EEG, we follow prior work~\cite{gmner} and evaluate performance using Precision (\textbf{Pre}), Recall (\textbf{Rec}), and the \textbf{F1} score. Each sample contains zero or more entity triples $\{(e_i,\ t_i,\ l_i) \}_{i=1}^{k_1}$, and we compute the correctness of each entity triple as follow:
\begin{equation}
correct =
\begin{cases}
1,& C_e \land C_t \land C_l,\\[2pt]
0,& \text{otherwise},
\end{cases}
\end{equation}
\begin{equation}
C_e/C_t =
\begin{cases}
1,& e_i/t_i=\hat{e_i}/\hat{t_i},\\[2pt]
0,& \text{otherwise},
\end{cases}
\end{equation}
\begin{equation}
C_l =
\begin{cases}
1,& l_i=\hat{l_i}=\text{None},\\[2pt]
1,& IoU(l_i, \hat{l_i}) \geq 0.5,\\[2pt]
0,& \text{otherwise},
\end{cases}
\end{equation}

\begin{equation}
IoU(l_i, \hat{l_i}) =
\frac{area(l_i \cap \hat{l_i})}
{area(l_i \cup \hat{l_i})},
\end{equation}
where $C_e$, $C_t$ and $C_l$ represent the correctness of entity, type and location; $e_i$, $t_i$ and $l_i$ represent the gold entity, type and location; $\hat{e_i}$, $\hat{t_i}$ and $\hat{l_i}$ represent the predicted entity, type and location; $IoU$ denotes the IoU score between $l_i$ and $\hat{t_i}$; and $area$ refers to the amount of two-dimensional space enclosed by a region. A predicted entity triple is regarded as correct only when the entity, type and location are all correct. Then Precision (Pre), Recall (Rec), and F1 score are used to evaluate the performance:
\begin{equation}
\text{Pre} = \frac{\#correct}{\#predict}, \quad
\text{Rec} = \frac{\#correct}{\#gold},
\end{equation}

\begin{equation}
\text{F1} = \frac{2 \times \text{Pre} \times \text{Rec}}{\text{Pre} + \text{Rec}},
\end{equation}
where $\#correct$, $\#predict$ and $\#gold$ respectively represent the number of triples of correct predictions and gold labels.
\paragraph{VG.}
Following prior work~\cite{grec}, we use \textbf{no-target accuracy} (\textbf{N-acc}) to measure localization accuracy when no target entity is present and Precision to measure accuracy when a target entity is present, and use Precision (Pre) to measure one-target entity localization. For a no-target sample, it considered a true positive (TP) when predicted bounding box is None, otherwise false negative (FN). Then N-acc is computed as follow:
\begin{equation}
\text{N-acc} = \frac{\text{TP}}{\text{TP} + \text{FN}}.
\end{equation}
For a one-target sample, a prediction is counted as a correct localization only if $IoU \ge 0.5$.
\paragraph{Quantitative metrics for textual bias.}
Inspired by N-acc, we introduce N-Pre, N-Rec, and N-F1 to quantify textual bias in GMNER, particularly cases where the model assigns bounding boxes to entities absent from the image. For an entity triple whose location is None ($l_i=\text{None}$), we determine its correctness as follows:
\begin{equation}
n\text{-}correct =
\begin{cases}
1,& C_e \land C_n,\\[2pt]
0,& \text{otherwise},
\end{cases}
\end{equation}

\begin{equation}
C_n =
\begin{cases}
1,& l_i=\hat{l_i}=\text{None},\\[2pt]
0,& \text{otherwise},
\end{cases}
\end{equation}
where $C_n$ represent the correctness of no-target entity location. We compute the above metrics over all no-target entity triples:
\begin{equation}
\text{N-Pre} = \frac{\#n\text{-}correct}{\#n\text{-}predict}, \quad
\text{N-Rec} = \frac{\#n\text{-}correct}{\#n\text{-}gold},
\end{equation}

\begin{equation}
\text{N-F1} = \frac{2 \times \text{N-Pre} \times \text{N-Rec}}{\text{N-Pre} + \text{N-Rec}},
\end{equation}
where $\#n\text{-}correct$, $\#n\text{-}predict$ and $\#n\text{-}gold$ respectively represent the number of triples of correct predictions and gold labels.
\paragraph{Quantitative metrics for visual bias.}
Directly inspecting every test image to quantify how MCR handles visual bias is impractical, so we introduce two indirect metrics that measure image-only entity recall. Based on whether a recalled entity appears in the input sentence, we define \textbf{N-Count} as the number of recalled entities that are absent from the sentence, and \textbf{N-Rate} as the proportion of such entities among all recalled entities. Specifically, for input sentence $s$ and the model predicted entity triples $\hat{\mathcal{Y}} = \{(\hat{e_i},\ \hat{t_i},\ \hat{l_i}) \}_{i=1}^{k_2}$ where $k_2$ is the number of all entity triples, we compute N-Count and N-Rate as follow:
\begin{equation}
\text{N-Count}=\frac{1}{k_2}\sum_{i}^{k_2}\mathbbm{1} \{\hat{e_i} \notin s\},
\end{equation}
\begin{equation}
\text{N-Rate}=\frac{\text{N-Count}}{k_2},
\end{equation} 
where $\mathbbm{1}$ denotes an indicator function that returns 1 if the predicted entity is mentioned in the sentence and 0 otherwise. A lower N-Count and N-Rate indicate weaker visual bias, as the model is less likely to hallucinate image-only entities as text mentions.

\subsection{Implementation Details}
\label{sec:implementation_detail}
We conduct all experiments on 8 NVIDIA Tesla L20 GPUs. Training and inference use the ms-swift~\cite{swift} framework, and decoding and sampling use the vLLM~\cite{vllm} engine. All training procedures are conducted using LoRA~\cite{lora}. 
\paragraph{MRSI.} we generate diverse reasoning schema using a combination of template-based extraction, DeepSeek~\cite{deepseek}, Qwen2.5VL-72B, and Qwen3VL-30B-A3B~\cite{qwen2_5vl}. We train Qwen2.5VL for 2 epochs and MimoVL for 5 epochs with a learning rate of $0.0001$ and cosine learning schedule. We process the data in batches of 16. We trained the model for 4 hours on 8 L20 GPUs.
\paragraph{CVO.} During the CVO phase, we train for 2 epochs with a learning rate of $0.000005$ and a batch size of 64. We use a warmup ratio of 0.05, sample 8 generations per input, set GRPO clipping thresholds to 0.15 and 0.25, and apply temperature 1.5 with top-$k$ sampling ($k=200$), top-$p$ sampling ($p=0.95$), and $\beta=0.005$. We trained the model for 20 hours on 8 L20 GPUs.

\subsection{Case Study}
\label{sec:case_study}
\begin{figure*}[ht]
  \centering
   \includegraphics[width=1.0\linewidth]{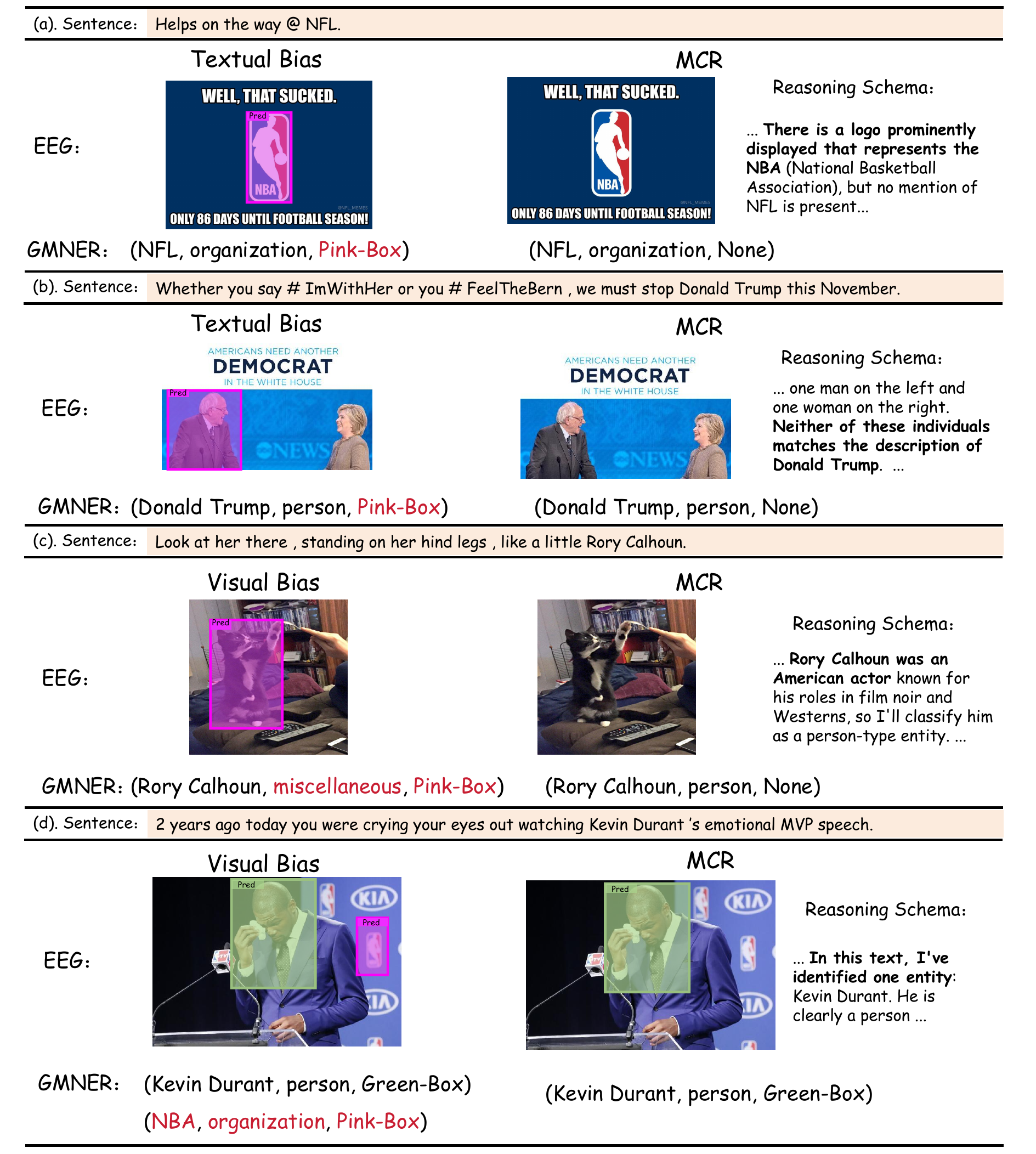}
   \caption{Case Studies of MCR Mitigating Modality Bias.}
   \label{fig:case_study_right}
\end{figure*}

\paragraph{MCR effectively mitigates modality bias.} As illustrated in Figure~\ref{fig:case_study_right}, (a) and (b) present two cases where MCR successfully mitigates textual bias. Naive End-to-end methods lead the model to assign incorrect image regions to textual entities. For example, the model incorrectly grounds the ``NBA'' logo to the textual entity ``NFL'', and assigns an elderly male to ``Donald Trump''. 

By explicitly generating reasoning paths and reinforcing cross-modal consistency verification, MCR effectively alleviates these issues. Specifically, the model recognizes the distinct semantics of the ``NBA'' logo in the image and the ``NFL'' entity in the text and correctly concludes that they do not match. Similarly, it explicitly verifies whether the person in the image corresponds to ``Donald Trump'', thereby avoiding erroneous grounding.

(c) and (d) present two cases where MCR successfully mitigates visual bias. When MLLMs perform entity recognition and classification, they can be distracted by irrelevant visual elements, leading to the spurious recall of image-only entities or incorrect classification of textual entities. For instance, the model erroneously recalls the image-only entity ``NBA'', and misclassifies the human entity ``Rory Calhoun'' due to the presence of a cat in the image.

By explicitly prompting the model to surface multimodal evidence and reinforcing the principled use of such evidence, MCR effectively alleviates these issues. Specifically, the model clarifies the modality source of each entity and relies on internal knowledge and sentence semantics when determining entity types, rather than being misled by superficial visual cues.

\begin{figure*}[ht]
  \centering
   \includegraphics[width=1.0\linewidth]{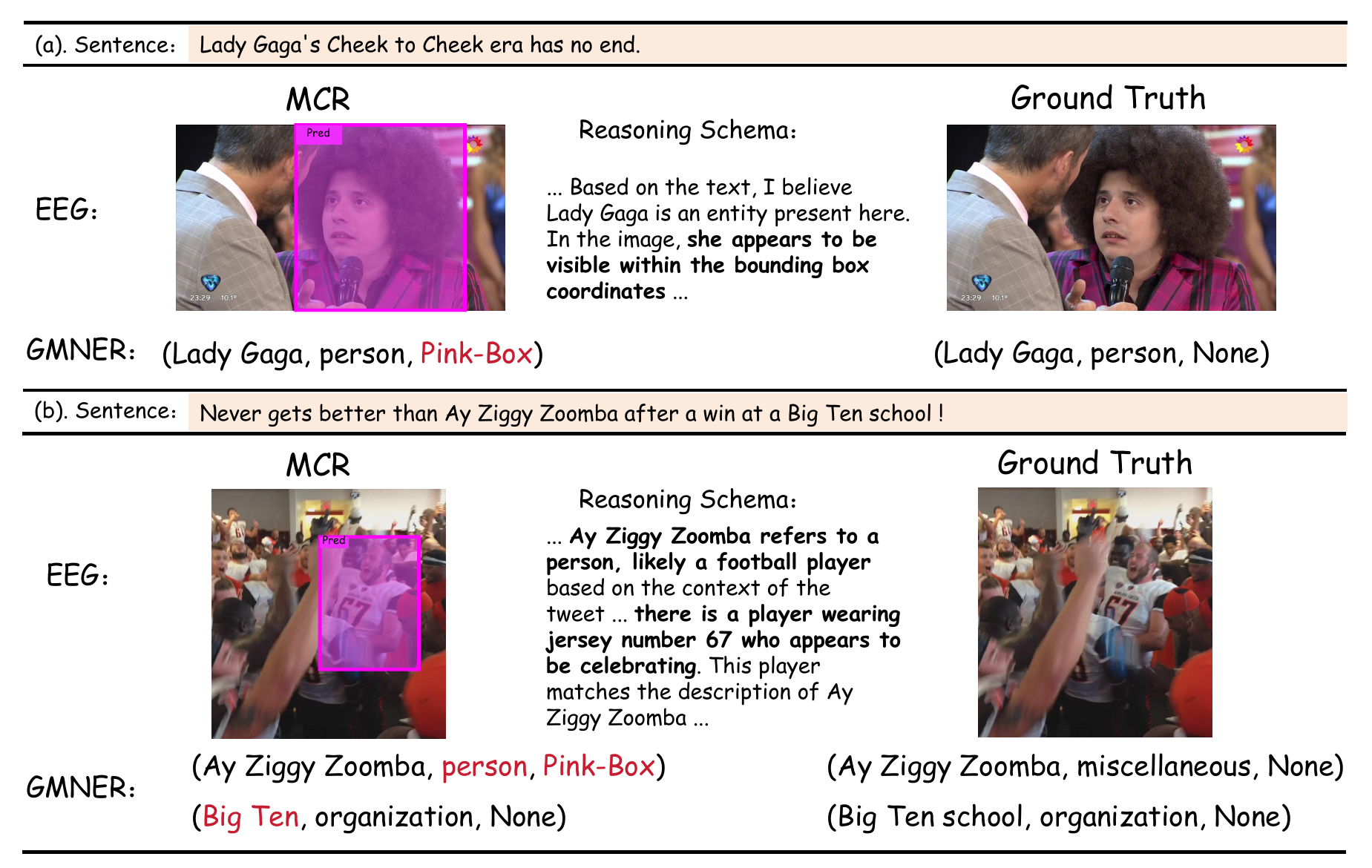}
   \caption{Failure Cases of MCR.}
   \label{fig:case_study_wrong}
\end{figure*}

\paragraph{Limitations of knowledge and entity span.} As illustrated in Figure~\ref{fig:case_study_wrong}, (a) and (b) present two failure cases of MCR. Although MLLMs incorporate substantial knowledge during training, GMNER requires broad, cross-domain knowledge that inevitably includes entities beyond the model’s coverage or cases where the model has acquired incorrect knowledge. As a result, MLLMs may still fail in such scenarios. For example, even though MCR possesses knowledge about ``Lady Gaga'', its limited visual knowledge of her appearance causes it to be misled by visually similar cues. Similarly, due to the lack of prior knowledge about ``Ay Ziggy Zoomba'', the model makes an incorrect judgment from the outset. These cases illustrate that MCR remains constrained by the underlying model’s knowledge coverage and visual familiarity. Figure~\ref{fig:case_study_wrong}(b) further shows that MLLMs still struggle with entity span detection.

\end{document}